\documentclass[letterpaper, 10 pt, conference]{cls/ieeeconf}
\IEEEoverridecommandlockouts
\let\labelindent\relax

\usepackage[normalem]{ulem}                         
\usepackage[table,usenames,dvipsnames]{xcolor}      
\usepackage{enumitem}                               
\usepackage{extarrows}                              
\usepackage[noadjust]{cite}                         
\usepackage{verbatim}

\usepackage{amsmath,amssymb,amsfonts,dsfont} 
\usepackage{amsthm}
\usepackage{bbm}
\usepackage{algorithm,algorithmicx,listings}        
\usepackage[noend]{algpseudocode}             

\usepackage{graphicx,tabularx,subcaption}
\usepackage[export]{adjustbox}
\usepackage{wrapfig}
\usepackage{makecell,booktabs}

\usepackage[font={small}]{caption}
\usepackage[titletoc,title]{appendix}

\usepackage[breaklinks=true, colorlinks, bookmarks=true, citecolor=Black, urlcolor=Violet,linkcolor=Black]{hyperref}

\usepackage{arydshln}               

\newcommand{\prl}[1]{\left(#1\right)}

\newcommand{\crl}[1]{\left\{#1\right\}}

\usepackage{mathtools}
\usepackage{pifont}
\usepackage{footnote}
\usepackage{tikz}
\usepackage{balance}
\makesavenoteenv{tabular}
\makesavenoteenv{table}

\def\etal/{et~al.}
\graphicspath{{fig/}}
\newcommand{\1}[1]{\mathds{1}_{#1}}

\newtheorem{theorem}{Theorem}
\newtheorem{proposition}{Proposition}

\theoremstyle{definition}
\newtheorem{definition}{Definition}
\newtheorem{assumption}{Assumption}
\newtheorem*{assumption*}{Assumption}
\newtheorem*{problem*}{Problem}
\newtheorem{problem}{Problem}
\theoremstyle{remark}

\newtheorem*{solution*}{Solution}

\def\thetitle{WFA-IRL: Inverse Reinforcement Learning of Autonomous Behaviors Encoded as Weighted Finite Automata}
\def\theauthor{Tianyu Wang and Nikolay Atanasov}
\def\thekeywords{keywords}

\hypersetup{
  pdfauthor={\theauthor},%
  pdftitle={\thetitle},%
  pdfsubject={\thetitle},%
  pdfkeywords={\thekeywords}
}

\newcommand{\calB}{{\cal B}}

\newcommand{\calD}{{\cal D}}

\newcommand{\calL}{{\cal L}}

\newcommand{\calP}{{\cal P}}

\newcommand{\calS}{{\cal S}}

\newcommand{\calU}{{\cal U}}

\newcommand{\calX}{{\cal X}}

\newcommand{\calAP}{{\cal AP}}

\newcommand{\bfm}{\mathbf{m}}

\newcommand{\bfp}{\mathbf{p}}

\newcommand{\bfs}{\mathbf{s}}

\newcommand{\bfu}{\mathbf{u}}

\newcommand{\bfx}{\mathbf{x}}

\newcommand{\bfalpha}{\boldsymbol{\alpha}}
\newcommand{\bfbeta}{\boldsymbol{\beta}}

\newcommand{\bftau}{\boldsymbol{\tau}}

\newcommand{\bfH}{\mathbf{H}}

\newcommand{\bfP}{\mathbf{P}}

\newcommand{\bfS}{\mathbf{S}}

\newcommand{\bfU}{\mathbf{U}}
\newcommand{\bfV}{\mathbf{V}}
\newcommand{\bfW}{\mathbf{W}}

\newcommand{\bfLambda}{\boldsymbol{\Lambda}}

\newcommand{\bbN}{\mathbb{N}}

\newcommand{\bbR}{\mathbb{R}}

\title{\LARGE \bf \thetitle}
\author{Tianyu Wang \and Nikolay Atanasov
\thanks{We gratefully acknowledge support from ONR SAI N00014-18-1-2828. The authors are with the Department of Electrical and Computer Engineering, University of California San Diego, La Jolla, CA 92093, USA {\tt\small \{tiw161,natanasov\}@eng.ucsd.edu}.}%
}

\begin{document}
\maketitle


\begin{abstract}
This paper presents a method for learning logical task specifications and cost functions from demonstrations. Constructing specifications by hand is challenging for complex objectives and constraints in autonomous systems. Instead, we consider demonstrated task executions, whose logic structure and transition costs need to be inferred by an autonomous agent. We employ a spectral learning approach to extract a weighted finite automaton (WFA), approximating the unknown task logic. Thereafter, we define a product between the WFA for high-level task guidance and a labeled Markov decision process for low-level control. An inverse reinforcement learning (IRL) problem is considered to learn a cost function by backpropagating the loss between agent and expert behaviors through the planning algorithm. Our proposed model, termed WFA-IRL, is capable of generalizing the execution of the inferred task specification in a suite of MiniGrid environments.
\end{abstract}

\section{Introduction}
\label{sec:introduction}

\begin{figure}[t]
\centering
\includegraphics[width=\linewidth]{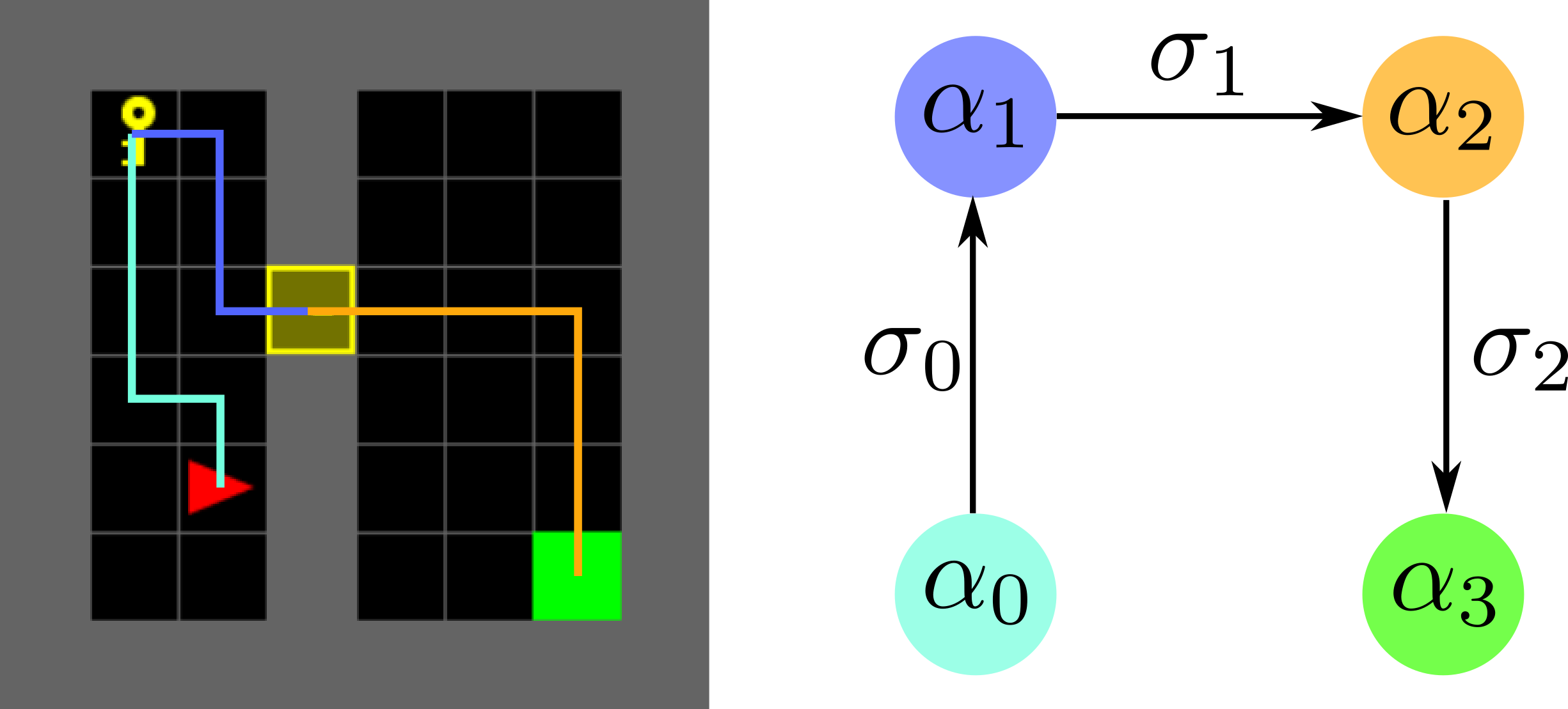}
\caption{(Left) An example trajectory in the MiniGrid environment~\cite{gym_minigrid}, where an agent has to pick up a key, open the door, and navigate to the goal. (Right) The trajectory can be decomposed into three segments, identified by hidden states $\bfalpha_t$. Transitions between the high-level states are triggered by events, such as picking up key ($\sigma_0$), or opening door ($\sigma_1$).}
\label{fig:doorkey_high_level}
\end{figure}

Autonomous systems are expected to achieve reliable performance in increasingly complex environments with increasingly complex objectives. Yet, it is often challenging to design a mathematical formulation that captures all safety and liveness requirements across various operational conditions. Minimizing a misspecified cost function may lead to undesirable performance, regardless of the quality of the optimization algorithm. However, a domain expert is often able to demonstrate desirable or undesirable behavior that implicitly captures the task specifications. As a simple illustration, consider the navigation task in Fig.~\ref{fig:doorkey_high_level}, requiring a door to be unlocked before reaching a goal state. Instead of encoding the task requirements as a cost function, an expert may provide several demonstrations of navigating to the goal, some of which require picking up the key whenever the door is locked. A reinforcement learning agent should infer the underlying logic sequence of the demonstrated task in order to learn the desired behavior.

Inverse reinforcement learning (IRL) \cite{Ng2000IRL, Ratliff2006MMP, Ziebart2008MaxEnt} focuses on inferring the latent costs of expert demonstrations. Early works assume that the cost is linear in a set of state features and minimize the feature expectation difference between learned policy and demonstrations. \cite{Ziebart2008MaxEnt} use dynamic programming to find a maximum entropy (MaxEnt) policy which maximizes the likelihood of the demonstrated actions. Later works \cite{Levine2011Nonlinear, Wulfmeier2016DeepMaxEnt} introduce Gaussian process or deep neural networks to learn nonlinear cost functions. \cite{Finn2016GCL} builds a connection between MaxEnt IRL and adversarial learning and solves continuous control problems. Most IRL models, however, consider general cost formulations that do not explicitly capture sequencing and compositional requirements of the demonstrated task \cite{Vazquez2017LearningTaskSpecs, Krishnan2019SWIRL}. Compared to a general cost formulation, this paper shows the logical structure of a complex task can be inferred from demonstrations. Exploiting the underlying task logic in planning ensures that the learned agent behavior mode matches the demonstrations.

Hierarchical reinforcement learning and options framework \cite{Sutton1999Temporal, Kulkarni2016HRL, Bacon2017Option, Riemer2018Learning} are formulations that learn task decomposition and temporal abstraction. Options are high-level macro-actions consisting of primitive actions. \cite{Fox2017DeepOptions} introduces a multi-level hierarchical model to discover options from demonstrations where option boundaries are inferred for trajectory segmentation. \cite{Kipf2019Compile} uses an unsupervised encoder-decoder model to predict subtask segmentation and categorical latent encoding. \cite{Xie2020DeepIL} uses graph recurrent neural networks with relational features between objects for high-level planning and low-level primitive dynamics prediction. 


Formal methods have been applied in robotics to prove and guarantee different behavioral properties such as safety and correctness \cite{Plaku2016Motion, Luckcuck2019FormalSA, Farrell2018Robotics, Fainekos2009LTL}. For example, linear temporal logic (LTL) \cite{Baier2008Principles} is used to specify safety and liveness objectives with temporal ordering constraints in control and reinforcement learning problems \cite{Kress2007Waldo, Kress2009LTL, Fainekos2009LTL, Fainekos2005TemporalLogic, Bhatia2010Sampling, Fu2016Optimal}. Specification mining of LTL formulas can learn finite state automata from execution traces \cite{Lemieux2015General, Kang2021Adversarial}. LTL formulas can also be inferred from Bayesian inference \cite{Shah2018Bayesian} or from graph connectivity of directed acyclic graphs over atomic propositions \cite{Chou2020LTL}. In this work, we consider weighted finite automata (WFA) in which the transitions carry weights. Whereas classical automata determine whether a word is accepted or rejected, WFA can compute quantitative values as a function of the weighted transitions from the execution of words \cite{Droste2009Handbook}. WFA offer the expressive power to model quantitative properties, such as resources, time or cost, of the demonstrated behavior. Under certain assumptions of the semiring on which the WFA is defined, it can be shown that WFA is expressively equivalant to weighted monadic second-order (MSO) logic \cite{Droste2005Weighted, Droste2009Handbook}.
We introduce an IRL model that learns to infer high-level task specifications and low-level control costs to imitate demonstrated behavior. Given a set of demonstrations, we use a spectral method to learn a WFA which encodes the task logic structure. The agent's interaction with the environment is modeled as a product between the learned WFA and a labeled Markov decision process (L-MDP). We propose a planning algorithm to search over the product space for a policy that satisfies task requirements encoded by the WFA. Since the true transition cost is not directly observable, we differentiate the error between the agent's policy and the demonstrated controls through the planning algorithm using a subgradient method introduced in \cite{Wang2020ICRA,Ratliff2006MMP}. We demonstrate that our WFA-IRL method correctly classifies accepting and rejecting sequences and learns a cost function that generalizes the demonstrated behavior to new settings in several MiniGrid environments \cite{gym_minigrid}. In summary, our \textbf{contribution} is to recognize that the logic structure of a demonstrated task can be learned as a weighted finite automaton and, in turn, can be integrated with differentiable task planning to learn generalizable behavior from demonstrations.



\section{Preliminaries}
\label{sec:preliminaries}

\subsection{Agent and environment models}
The agent's interaction with the environment is modeled as an L-MDP \cite{Ding2011MDP}.

\begin{definition}
A labeled Markov decision process is a tuple $\{\calX, \calU, \bfx_0, f, c, \calAP, \ell\}$, where $\calX$, $\calU$ are finite sets of states and controls, $\bfx_0 \in \calX$ is an initial state, $f: \calX \times \calU \rightarrow \calX$ is a deterministic transition function, and $c: \calX \times \calU \rightarrow \mathbb{R}_{\geq 0}$ assigns a non-negative cost when control $\bfu \in \calU$ is applied at state $\bfx \in \calX$. A finite set of atomic propositions $\calAP$ provides logic statements that must be true or false (e.g., ``the agent is 1 meter away from the closest obstacle'' or ``the agent possesses a key''). A labeling function $\ell: \calX \times \calU \rightarrow 2^{\calAP}$ assigns a set of atomic propositions that evaluate true for a given state transition.
\end{definition}

We assume that the state $\bfx$ is fully observable and captures both endogenous variables for the agent, such as position and orientation, and exogenous variables, such as an environment containing objects of interest as illustrated in Fig.~\ref{fig:doorkey_high_level}. The transition function $f(\bfx,\bfu)$ specifies the change of state $\bfx$ when control $\bfu$ is executed, and $c(\bfx,\bfu)$ assigns a non-negative cost to this transition. The alphabet of the L-MDP is the set of labels $\Sigma = 2^{\calAP}$ that can be assigned to the transitions. The labeling function $\sigma = \ell(\bfx,\bfu)$ provides the atomic propositions $\sigma \in \Sigma$ which are satisfied during the transition $f(\bfx,\bfu)$. The set of words on $\Sigma$ is denoted by $\Sigma^*$ and consists of all strings $\sigma_{0:T} = \sigma_0\dots\sigma_T$ for $\sigma_t \in \Sigma$ and $T \in \bbN$. We assume that the transition $f$ and labeling $\ell$ are known. However, the cost function $c$ is unknown and needs to be inferred from expert demonstrations.



\subsection{Expert model}
\label{sec:expert_model}
%
%
The agent needs to execute a task, whose success is evaluated based on the word $\sigma_{0:T} \in \Sigma^*$ resulting from the agent's actions. We model the quality of the task execution by a function $h: \Sigma^* \rightarrow \bbR$. An execution $\sigma_{0:T}$ is deemed successful if $h(\sigma_{0:T}) \geq \xi$ for a known performance threshold $\xi$, and unsuccessful otherwise. As argued in the introduction, defining the function $h$ explicitly is challenging in many applications. Instead, we consider a training set $\calD = \crl{(\bfx_{0:T_n}^n, \bfu_{0:T_n}^n, s^n)}_{n=1}^N$ of $N$ demonstrations of the same task in different environment configurations provided by an expert. Each demonstration $n$ contains the controls $\bfu_{0:T_n}^n = \bfu_0^n\dots\bfu_{T_n}^n$ executed by the expert, the resulting agent-environment states $\bfx_{0:T_n}^n = \bfx_0^n\dots\bfx_{T_n}^n$, and the success level $s^n\in \bbR$ of the execution, measured by $h(\sigma_{0:T_n}^n)$, where $\sigma_t^n = \ell(\bfx_t^n,\bfu_t^n)$ is the label encountered by the expert at time $t$. We assume that the expert knows the \emph{true} task $h$ and the \emph{true} cost $c$ and can solve a finite-horizon first-exit deterministic optimal control problem \cite{Bertsekas1995OC} over the L-MDP:
\begin{equation}
\label{eq:Q_star}
\begin{aligned}
&Q^*(\bfx, \bfu) := \min_{T,\bfu_{1:T}} \sum_{t=0}^{T} c(\bfx_t, \bfu_t) \\
&\;\;\text{s.t.}\;\; \bfx_{t+1} = f(\bfx_t, \bfu_t),\; \bfx_0 = \bfx,\; \bfu_0 = \bfu, \\
&\;\;\;\; \sigma_t = \ell(\bfx_t,\bfu_t),\; h(\sigma_{0:T}) \geq \xi,
\end{aligned}
\end{equation}
%
where $Q^*(\bfx, \bfu)$ is the optimal value function. Since \eqref{eq:Q_star} is a deterministic optimal control problem, there exists an open-loop control sequence which is optimal, i.e., achieves the same cost as an optimal closed-loop policy function \cite[Chapter 6]{Bertsekas1995OC}. However, we consider experts that do not necessarily choose strictly rational controls. Instead, we model the expert behavior using a stochastic Boltzmann policy over the optimal values $\pi^*(\bfu | \bfx) \propto \exp\prl{-\frac{1}{\eta} Q^*(\bfx, \bfu)}$,
where $\eta \in (0, \infty)$ is a temperature parameter representing a continuous spectrum of rationality. For example, $\eta \rightarrow 0$ means that the expert takes strictly optimal controls while $\eta \rightarrow \infty$ means random controls are selected. The Boltzmann expert model was previously introduced and studied in \cite{Neu2012Apprenticeship, Ramachandran2007BayesianIRL, Wang2020ICRA}. It provides an exponential preference for controls that incur low long-term costs. This expert model also allows efficient policy search, as shown in Sec.~\ref{sec:planning}, and computation of the policy gradient with respect to the cost needed to optimize the cost parameters, as shown in Sec.~\ref{sec:cost_learning}.


\section{Problem Statement}
\label{sec:problem_statement}


The agent needs to infer the unknown task model $h$ and unknown cost function $c$ from the expert demonstrations $\calD = \crl{(\bfx_{0:T_n}^n, \bfu_{0:T_n}^n, s^n)}_{n=1}^N$.

\begin{problem}
\label{pb:h}
Given the demonstrations $\calD$ and labeling $\sigma_t^n = \ell(\bfx_t^n,\bfu_t^n)$, optimize the parameters $\psi$ of an approximation $h_\psi$ of the unknown task function $h$ to minimize the mean squared error:
\begin{equation}
\label{eq:wfa_loss}
\min_\psi \calL_h(\psi) :=\frac{1}{N}\sum_{n=1}^N \prl{h_\psi(\sigma_{0:T_n}^n) - s^n}^2.
\end{equation}
\end{problem}

Similarly, the agent needs to obtain an approximation $c_\theta$ with parameters $\theta$ of the unknown cost function $c$. This allows the agent to obtain a control policy:
\begin{equation}
\label{eq:boltzmann_agent_policy}
  \pi_\theta(\bfu | \bfx) \propto \exp\prl{-\frac{1}{\eta} Q_\theta(\bfx, \bfu)},
\end{equation}
approximating the expert model using a value function $Q_\theta$ computed according to \eqref{eq:Q_star} with $c$ and $h$ replaced by $c_\theta$ and $h_\psi$, respectively.


\begin{problem}
Given the demonstrations $\calD$, optimize the parameters $\theta$ of an approximation $c_\theta$ of the unknown cost function $c$ such that the log-likelihood of the demonstrated controls $\bfu_t^n$ is maximized under the agent policy in \eqref{eq:boltzmann_agent_policy}:
\begin{equation}
\label{eq:loss}
\min_\theta \calL_c(\theta) := -\sum_{n=1}^N \1{\crl{s^n \geq \xi}} \sum_{t=0}^{T_n} \log \pi_\theta(\bfu_t^n|\bfx_t^n),
\end{equation}
where $\1{}$ is an indicator function and $\xi$ is the known task satisfaction threshold.
\end{problem}

\section{Technical Approach}
\label{sec:technical_approach}

We first discuss how to learn a task model $h_\psi$ from demonstrations $\calD$ in Sec.~\ref{sec:spectral_method}. Next, in Sec.~\ref{sec:planning}, we learn a cost model $c_{\theta}$ by solving the optimal control problem in \eqref{eq:Q_star} to obtain an agent policy $\pi_\theta$. Finally, in Sec.~\ref{sec:cost_learning}, we show how to backpropagate the policy loss $\calL_c(\theta)$ in \eqref{eq:loss} through the optimal control problem to update the cost parameters $\theta$.


\subsection{Spectral learning of task specifications}
\label{sec:spectral_method}

\begin{figure}[t]
\centering
\includegraphics[width=\linewidth]{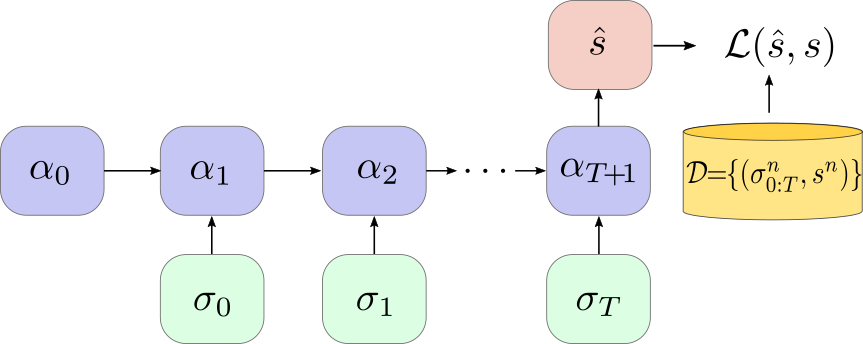}
\caption{Inferring the hidden state progression $\bfalpha_t$ from events $\sigma_t$ can be acheived by an RNN with initial hidden state $\bfalpha_0$, hidden state transition $\bfalpha_{t+1} = g_1(\sigma_t, \bfalpha_t, \bfW)$ and output $\hat{s} = h_\psi(\sigma_{0:T})= g_2(\bfalpha_{T\!+\!1}, \bfbeta)$, where $g_1, g_2$ are nonlinear functions. The weights $\psi = \prl{\bfalpha_0, \bfW, \bfbeta}$ can be learned via the loss $\calL(\hat{s}, s)$ in \eqref{eq:wfa_loss} between RNN outputs $\hat{s}$ and demonstration scores $s$.}
\label{fig:rnn}
\end{figure}

Fitting a single cost neural network $c_\theta$ that is capable of generalizing to various environment configurations and tasks is difficult when state and control spaces are large and the task horizon is long. An alternative is to consider the cost function and its corresponding policy only for small segments of the task, associated with different subtasks. This idea is based on the observation that task specifications often have a compositional logic structure. For example, the demonstrated trajectory in the DoorKey environment in Fig.~\ref{fig:doorkey_high_level} can be decomposed into three segments, each denoted by a high-level state $\bfalpha$. The transitions between the high-level states are triggered by events like $\sigma_0$: a key is picked up, and $\sigma_1$: a door is opened. Note that there is no direct transition between $\bfalpha_1$ and $\bfalpha_3$ because the door cannot be opened without possessing a key. Such high-level state abstraction and transitions are commonly learned via recurrent neural network (RNN) or memory architectures \cite{Hausknecht2015DRQN, Mirowski2016Learning}. For example, to solve Problem~\ref{pb:h}, we can use an RNN $h_\psi$ in Fig.~\ref{fig:rnn} with initial hidden state $\bfalpha_0$, hidden state transition $\bfalpha_{t+1} = g_1(\sigma_t, \bfalpha_t, \bfW)$ and output function $h_\psi(\sigma_{0:T}) = g_2(\bfalpha_{T\!+\!1}, \bfbeta)$, where $g_1, g_2$ are nonlinear functions and $\psi = \prl{\bfalpha_0, \bfW, \bfbeta}$ are learnable weights. Instead of an RNN model, in this work, we propose to use a weighted finite automaton (WFA) \cite{Balle2012Spectral} to represent $h_\psi$. A WFA is less expressive than an RNN \cite{Rabusseau2019WFA_2RNN} but can be trained more effectively from small demonstration dataset. Moreover, a WFA generalizes deterministic and nondeterministic finite automata, which are commonly used to model logic task specifications for autonomous agents \cite{Kress2009LTL,Fainekos2009LTL,Kress2007Waldo,Fainekos2005TemporalLogic}. Hence, a WFA model is sufficiently expressive to represent a complex task and allows one to focus on a temporal abstraction without reliance on the low-level system dynamics.

\begin{definition}
A weighted finite automaton (WFA) with $m$ states is a tuple $\psi = \crl{\bfalpha_0, \bfbeta, \crl{\bfW_\sigma}_{\sigma \in \Sigma}}$ where $\bfalpha_0, \bfbeta \in \bbR^m$ are initial and final weight vectors and $\bfW_{\sigma} \in \bbR^{m\times m}$ are transition matrices associated with each symbol $\sigma\in \Sigma$. A WFA $\psi$ represents a function $h_\psi: \Sigma^* \rightarrow \mathbb{R}$ by $h_\psi(\sigma_{0:T}) = \bfalpha_0 ^\top \bfW_{\sigma_0} \bfW_{\sigma_1} \dots \bfW_{\sigma_T} \bfbeta$.
\end{definition}

A WFA represents the task progress for a given word $\sigma_{0:t}$ via $h_\psi(\sigma_{0:t}) = \bfalpha_0 ^\top \bfW_{\sigma_0} \bfW_{\sigma_1} \dots \allowbreak\bfW_{\sigma_t} \bfbeta$, where the high-level state at time $t\!+\!1$ is $\bfalpha_{t\!+\!1} = \prl{\bfalpha_0 ^\top \bfW_{\sigma_0} \bfW_{\sigma_1} \dots \bfW_{\sigma_t}}^\top$. When the WFA is learned correctly, its prediction for an expert word should approximate the expert score $s$, i.e., $h_\psi(\sigma_{0:T}) \approx s$. This can be used to guide a task planning algorithm by providing a task satisfaction criterion. A trajectory with corresponding word $\sigma_{0:T}$ is identified as successful if the WFA prediction passes the known performance threshold introduced in Sec.~\ref{sec:expert_model}, i.e., $h_\psi(\sigma_{0:T}) = \bfalpha_{T\!+\!1}^\top\bfbeta \geq \xi$.

Our approach to learn a minimal WFA is based on the spectral learning method developed by \cite{Balle2012Spectral}. The spectral method makes use of a Hankel matrix $\bfH_h \in \mathbb{R}^{\Sigma^* \times \Sigma^*}$ associated with the function $h: \Sigma^* \rightarrow \mathbb{R}$, which is a bi-infinite matrix with entries $\bfH_h(u, v) = h(uv)$ for $u, v \in \Sigma^*$. We assume the class of functions $h$ that can be represented by a WFA are rational power series functions and their associated Hankel matrix $\bfH_h$ has finite rank \cite{Berstel1988Rational, Salomaa2012Automata}. 
It can be showns that under certain assumptions WFA are expressively equivalent to monadic second-order logic. The quantitative property of WFA allows us to model the performance score $s$ of the demonstrated trajectories.


\begin{assumption}
The Hankel matrix $\bfH_{h}$ associated with the true task specification $h$ has finite rank.
\end{assumption}

In practice, only finite sub-blocks of the Hankel matrix, constructed from the expert demonstrations $\calD$, can be considered. Given a basis $\calB = (\calP, \calS)$ where $\calP, \calS \subset \Sigma^*$ are finite sets of prefixes and suffixes respectively, define $\bfH_{\calB}$ and $\{\bfH_{\sigma}\}_{\sigma \in \Sigma}$ as the finite sub-blocks of $\bfH_h$ such that $\bfH_{\calB}(u, v) = h(uv)$, $\bfH_\sigma(u, v) = h(u\sigma v)$, $\forall u \in \calP, v \in \calS$. The foundation of the spectral learning method is summarized in the following theorem.

\begin{theorem}[\cite{Balle2012Spectral}]
\label{th:spectral}
Given a basis $\calB = (\calP, \calS)$ such that the empty string $\lambda \in \calP \cap \calS$ and $rank(\bfH_h) = rank(\bfH_{\calB})$, for any rank $m$ factorization $\bfH_{\calB} = \bfP\bfS$ where $\bfP \in \mathbb{R}^{|\calP| \times m}$ and $\bfS \in \mathbb{R}^{m \times |\calS|}$, the WFA $\{\bfalpha_0, \bfbeta, \{\bfW_{\sigma}\}\}$ is a minimal WFA representing $h$, where $\bfalpha_0^\top = \bfP(\lambda, :)$ is the row vector of $\bfP$ corresponding to prefix $\lambda$, $\bfbeta = \bfS(:, \lambda)$ is the column vector of $\bfS$ corresponding to suffix $\lambda$, and $\bfW_{\sigma} = \bfP^\dagger \bfH_{\sigma} \bfS^\dagger$, $\forall \sigma \in \Sigma$.
\end{theorem}

A basis can be chosen empirically from demonstrations $\calD$. For example, we can choose a basis that includes all prefixes and suffixes that appear in the words $\crl{\sigma_{0:T_n}^n}$ or one with desired cardinality for the most frequent prefixes and suffixes. Given a basis, the Hankel blocks $\bfH_{\calB}$, $\crl{\bfH_{\sigma}}$ are constructed from $\calD$. For example, given a word and its score $\prl{\sigma_{0:T}, s}$, we set the entries $\bfH_{\calB}(\lambda, \sigma_{0:T})$, $\bfH_{\calB}(\sigma_0, \sigma_{1:T})$, $\dots$, $\bfH_{\calB}(\sigma_{0:T}, \lambda)$, and $\bfH_{\sigma}(\sigma_{0:t-1}, \sigma_{t+1: T})$, where $\sigma = \sigma_t$ with value $s$. To find a low rank factorization of $\bfH_{\calB}$, we use truncated singular value decomposition, $\bfH_{\calB} = \bfU_m\bfLambda_m\bfV_m^\top$ where $\bfLambda_m$ is a diagonal matrix of the $m$ largest singular values and $\bfU_m, \bfV_m$ are the corresponding column vectors, and set $\bfP=\bfU_m$ and $\bfS = \bfLambda_m\bfV_m^\top$. Finally, the vectors and matrices $\psi = \{\bfalpha_0, \bfbeta, \crl{\bfW_{\sigma}}\}$ of the WFA can be obtained from $\bfP$, $\bfS$, $\crl{\bfH_\sigma}$ using Theorem~\ref{th:spectral}. 
\subsection{Planning in a product WFA-MDP system}
\label{sec:planning}



Given a learned WFA representation $h_{\psi}$ and an initial cost estimate $c_{\theta}$, we propose a planning algorithm to solve the deterministic optimal control problem in \eqref{eq:Q_star} and obtain a control policy $\pi_\theta(\bfu | \bfx)$ as in \eqref{eq:boltzmann_agent_policy}. To determine the termination condition for the problem in \eqref{eq:Q_star}, we define the product of the WFA, modeling the task, and the L-MDP, modeling the agent-environment interactions.

\begin{definition}
Given an L-MDP $\{\calX, \calU, \bfx_0, f, c, \calAP, \ell\}$ and a WFA $\{\bfalpha_0, \bfbeta, \crl{\bfW_{\sigma}}\}$, a product WFA-MDP model is a tuple $\crl{\calS, \calU, \bfs_0, T, \calS_F, c, \calAP, \ell}$ where $\calS = \calX \times \bbR^m$ is the product state space, $\bfs_0 = (\bfx_0, \bfalpha_0)$ is the initial state, and $\calS_F = \crl{(\bfx, \bfalpha) \in \calS \mid \bfalpha^\top \bfbeta \geq \xi}$ are the final states. The function $T: \calS \times \calU \rightarrow \calS$ is a deterministic transition function such that $T((\bfx_t, \bfalpha_t), \bfu_t) = (\bfx_{t+1}, \bfalpha_{t+1})$ where $\bfx_{t+1} = f(\bfx_t, \bfu_t)$, emitting symbol $\sigma_{t} = \ell(\bfx_t, \bfu_t)$ and causing transition $\bfalpha_{t+1} = \bfW_{\sigma_t}^\top \bfalpha_t$.
\end{definition}

To obtain the agent policy in \eqref{eq:boltzmann_agent_policy} for any state $\bfx_t \in \calX$ and control $\bfu_t \in \calU$, our goal is to compute the optimal cost-to-go values for the WFA-MDP model:
\begin{align}
\label{eq:Q_V}
Q_\theta(\bfs_t, \bfu_t) &= c_\theta(\bfx_t, \bfu_t) + V_\theta(T(\bfs_t, \bfu_t)) \notag\\
&= c_\theta(\bfx_t, \bfu_t) + \min_{T, \bfu_{t+1:T}} \sum_{k=t+1}^T c_\theta(\bfx_k, \bfu_k)
\end{align}
where $\bfs_{t+1} = T(\bfs_t, \bfu_t)$ and $\bfalpha_{T\!+\!1}^\top \bfbeta \geq \xi$. We have rewritten the terminal state condition as $\bfalpha_{T\!+\!1}^\top \bfbeta \geq \xi$, where we keep track of the task hidden state $\bfalpha_t$ using the WFA-MDP transition function $T$. Our key observation is that \eqref{eq:Q_V} is a deterministic shortest path problem and $V_\theta(T(\bfs_t, \bfu_t))$ can be obtained via any shortest path algorithm, such as Dijkstra \cite{Dijkstra1959Dijkstra}, A* \cite{ARA} or RRT* \cite{Karaman_RRTstar11}. When we use a shortest path algorithm to update the cost-to-go values of successor states $\bfs_{t+1} = T(\bfs_t, \bfu_t)$, we concurrently compute the corresponding WFA state $\bfalpha_{t+1} = \bfW_{\sigma_t}^\top \bfalpha_t$ where $\sigma_t = \ell(\bfx_t, \bfu_t)$. A goal state $\bfs_{T\!+\!1}$ is reached when its WFA state $\bfalpha_{T\!+\!1}$ satisfies $\bfalpha_{T\!+\!1}^\top \bfbeta \geq \xi$.
The agent policy $\pi_\theta$ in~\eqref{eq:boltzmann_agent_policy} with respect to the current cost estimate $c_\theta$ can be obtained from the cost-to-go values $Q_\theta$ in \eqref{eq:Q_V} computed by the shortest path algorithm.

\subsection{Optimizing cost parameters}
\label{sec:cost_learning}

We discuss how to differentiate the loss function $\calL_c(\theta)$ in \eqref{eq:loss} with respect to $\theta$ through the deterministic shortest path problem defined by the product WFA-MDP model. \cite{Wang2020ICRA} introduce a sub-gradient descent approach to differentiate the log likelihood of expert demonstrations from the Bolzman policy in~\eqref{eq:boltzmann_agent_policy} through the optimal cost-to-go values in \eqref{eq:Q_V}. The cost parameters can be updated by (stochastic) subgradient descent at each iteration $k$ with learning rate $\gamma^{(k)}$, $\theta^{(k+1)} = \theta^{(k)} - \gamma^{(k)} \nabla\calL_c(\theta^{(k)})$. Intuitively, the subgradient descent makes the trajectory starting with a demonstrated control more likely, while those with other controls less likely. The analytic subgradient computation is presented below.

\begin{proposition}\cite[Proposition~1]{Wang2020ICRA}
\label{prop:chain_rule}
Consider an expert transition $\prl{\bfx_t, \bfu_t}$. Define $\bftau(\bfx_t, \bfu)$ as the optimal path starting from state $\bfx_t$ and any control $\bfu \in \calU$ that achieves $Q_\theta(\bfx_t, \bfu)$ in \eqref{eq:Q_V} under cost estimate $c_\theta$. A subgradient of the agent policy \eqref{eq:boltzmann_agent_policy} evaluated at expert transition $\prl{\bfx_t, \bfu_t}$ with respect to cost parameters $\theta$ can be obtained via the chain rule as:
\begin{align}
\label{eq:subgradient}
&\frac{\partial \log \pi_\theta(\bfu_t \mid \bfx_t)}{\partial \theta} 
= \sum_{\bfu' \in \calU} \frac{d \log \pi_\theta(\bfu_t \mid \bfx_t)}{d Q_\theta(\bfx_t, \bfu')} \frac{\partial Q_\theta(\bfx_t, \bfu')}{\partial \theta} \notag \\
&\;\;= \sum_{\bfu' \in \calU} \frac{1}{\eta} \prl{\1{\crl{\bfu' = \bfu_t}} - \pi_\theta(\bfu_t \mid \bfx_t)} \notag\\
&\;\;\;\;\times\sum_{\prl{\bfx, \bfu}\in \bftau(\bfx_t, \bfu')} \frac{\partial Q_\theta(\bfx_t, \bfu')}{\partial c_\theta(\bfx, \bfu)} \frac{\partial c_\theta(\bfx, \bfu)}{\partial \theta}
\end{align}
\end{proposition}

Substituting \eqref{eq:subgradient} in the gradient of $\calL_c(\theta)$ in \eqref{eq:loss}, Proposition~\ref{prop:chain_rule} provides an explicit subgradient computation to allow backpropagation with respect to $\theta$ through the value function $Q_\theta$ of the deterministic shortest path problem in \eqref{eq:Q_V}. The subgradient only affects the cost parameters through the optimal trajectories $\bftau(\bfx_t, \bfu'),\, \forall \bfu' \in \calU$ for expert transitions $(\bfx_t, \bfu_t)$ which can be retrieved from any optimal planning algorithm applied in Sec~\ref{sec:planning}. Thereafter, the cost parameters can be optimized depending on the specific form of $\frac{\partial c_\theta(\bfx,\bfu)}{\partial \theta}$. 

\begin{figure}[t]
\centering
\includegraphics[width=\linewidth]{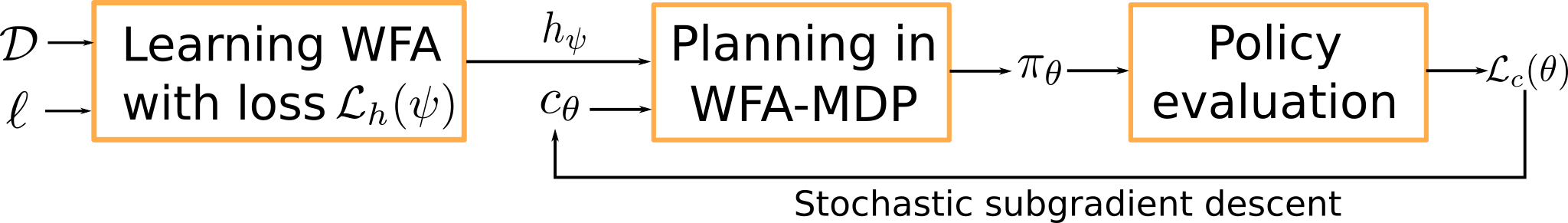}
\caption{WFA-IRL architecture for joint learning of a task specification $h_{\psi}$ and cost function $c_{\theta}$. Given demonstrations $\calD$ and a labeling function $\ell$, we learn the unknown task specification with a weighted finite automaton. We construct a product WFA-MDP space from the learned WFA $\psi=\crl{\bfalpha_0, \bfbeta, \crl{\bfW_\sigma}_{\sigma\in \Sigma}}$ to solve a deterministic shortest path problem with cost estimate $c_\theta$. The agent policy $\pi_\theta$ is compared with the demonstrated controls to backpropagate the loss $\calL_c(\theta)$ with respect to $\theta$.}
\label{fig:approach}
\end{figure}

Our complete approach WFA-IRL is summarized in Fig.~\ref{fig:approach}. We first solve Problem~\ref{pb:h} to find a WFA $h_\psi$ which models the demonstrated task. The learned WFA provides termination conditions for a deterministic shortest path problem in the product WFA-MDP. Cost parameters are optimized by backpropagating the loss in \eqref{eq:loss} through the planning algorithm.

\section{Evaluation}
\label{sec:experiments}

\begin{figure}[t]
\centering
\includegraphics[width=\linewidth]{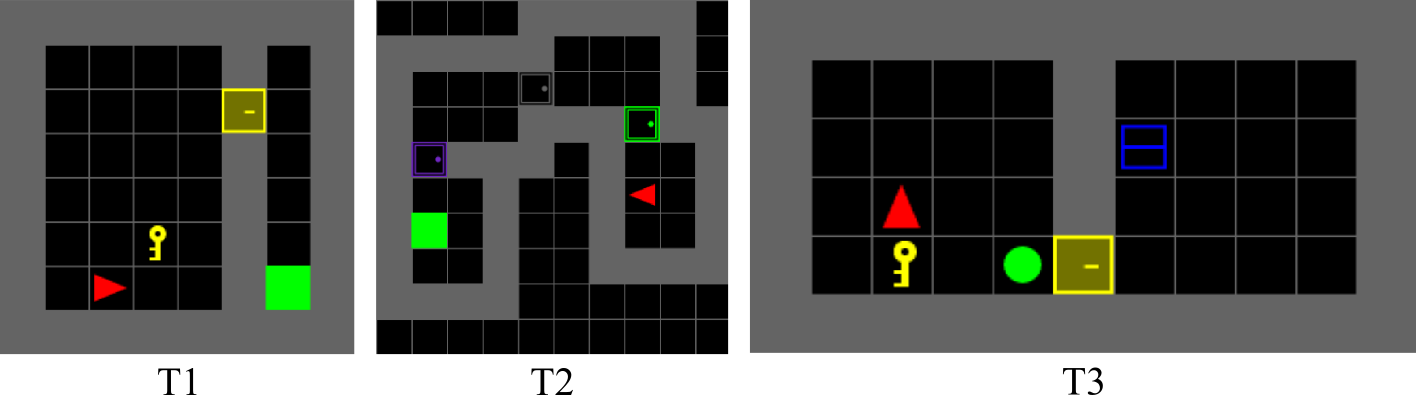}
\caption{MiniGrid environments \cite{gym_minigrid} of our experiments. In T1 (\textit{MiniGrid-MultiRoom-N4-S5-v0}) the agent must pick up $Key$ to unlock $Door$ and reach $Goal$ in the other room. In T2 (\textit{MiniGrid-MultiRoom-N4-S5-v0}) it has to open a series of $Door$s to reach $Goal$ in the last room. In T3 (\textit{MiniGrid-BlockedUnlockPickup-v0}) it has to move away a blocking $Ball$, unlock $Door$ with $Key$ and pick up $Box$. The state $\bfx_t$ includes the grid image $\bfm_t \in \crl{Wall, Key, Door, Box, Ball, Empty}^{H \times W}$, the agent position $\bfp_t \in \crl{1,\ldots,H}\times\crl{1,\ldots,W}$, direction $d_t \in \crl{Up, Left, Down, Right}$, and the object carried $o_t \in \crl{Key, Ball, Box, Empty}$. The control space $\calU$ is defined as turn left/right, move forward, pick up/drop/toggle an object.}
\label{fig:minigrid_envs}
\end{figure}

We consider three MiniGrid tasks shown in Fig.~\ref{fig:minigrid_envs} whose atomic propositions are shown in Table~\ref{tb:atomic_propositions}. The task specifications can be expressed in terms of these propositions, e.g., one possible trajectory that fulfills task T1 is to evaluate the propositions $p_1, p_2, p_3$ as true sequentially.

\begin{table}[ht]
\renewcommand*{\arraystretch}{1.4}
\centering
\caption{Atomic propositions used in each task.}
\begin{tabular}{|m{1em} | m{2.3cm}| m{2.3cm}| m{2cm} |} 
  \hline
    & T1 & T2 & T3 \\ 
  \hline
  $p_1$ & $Key$ is picked up & $Door$ 1 is open & \makecell{$Ball$ is 2 steps\\away from $Door$} \\ 
  \hline
  $p_2$ & $Door$ is open & $Door$ 2 is open & $Key$ is picked up \\
  \hline
  $p_3$ & Agent reaches $Goal$ & $Door$ 3 is open & $Door$ is open \\
  \hline
  $p_4$ & ------ & Agent reaches $Goal$ & $Box$ is picked up \\
  \hline
\end{tabular}
\label{tb:atomic_propositions}
\end{table}

\subsection{Demonstrations}
\label{sec:demonstrations}
An expert trajectory is collected by iteratively rolling out the controls sampled from the expert policy $\pi^*$ at each state $\bfx$, where $Q^*(\bfx, \bfu)$ in \eqref{eq:Q_star} is computed via the Dijkstra's algorithm with cost of $1$ for any feasible transition. For each task, we consider two sets of expert demonstrations $\textcolor{OliveGreen}{\calD_1}$ and $\textcolor{Orange}{\calD_2}$, each with 32 trajectories collected from expert policies with temperatures $\eta\in\{0, 0.5\}$. The expert trajectories in $\textcolor{OliveGreen}{\calD_1}$ and $\textcolor{Orange}{\calD_2}$ are strictly optimal and suboptimal, respectively, and are labeled with score $s=1$. In each set we also add $128$ failed trajectories with score $s=0$ from a random exploration policy (effectively setting expert policy temperature $\eta \rightarrow \infty$). The full demonstration set is used in each case to learn a WFA representation $h_\psi$ of the task via the spectral method in Sec.~\ref{sec:spectral_method} while only the successful trajectories are used to learn the cost function $c_\theta$, as in Sec.~\ref{sec:planning} and~\ref{sec:cost_learning}.

\begin{table}[t]
\centering
\caption{Results on MiniGrid environment tasks. In each entry, \textcolor{OliveGreen}{Green} / \textcolor{Orange}{Orange} are results trained from demonstrations $\color{OliveGreen}{\calD_1}$ with expert policy temperature $\eta=0$ and $\color{Orange}{\calD_2}$ with $\eta=0.5$, respectively. Top: Best Scikit-SpLearn hyperparameters that solve Problem~\ref{pb:h} for each task. Bottom: Mean episode returns (or negative cumulative true cost, higher is better) are reported across 64 randomly generated test environments.}
\label{tb:results}
\begin{tabular}{|c | c | c | c|}
    \hline
      & T1 & T2 & T3 \\ 
    \hline
    $\texttt{rank}$ & $\color{OliveGreen}{5} / \color{Orange}{9}$  & $\color{OliveGreen}{6} / \color{Orange}{9}$  & $\color{OliveGreen}{7} / \color{Orange}{11}$  \\ 
    \hline
    $\texttt{rows}$ & $\color{OliveGreen}{4} / \color{Orange}{5}$  & $\color{OliveGreen}{5} / \color{Orange}{5}$  & $\color{OliveGreen}{6} / \color{Orange}{6}$  \\
    \hline
    $\texttt{cols}$ & $\color{OliveGreen}{4} / \color{Orange}{5}$  & $\color{OliveGreen}{5} / \color{Orange}{6}$  & $\color{OliveGreen}{6} / \color{Orange}{7}$  \\
    \hline
\end{tabular}
\begin{tabular}{|c | c | c | c|}
  \hline
      & T1 & T2 & T3 \\ 
  \hline
  BC & $\color{OliveGreen}{0.364} / \color{Orange}{0.253}$ & $\color{OliveGreen}{0.338} / \color{Orange}{0.307}$ & $\color{OliveGreen}{0.284} / \color{Orange}{0.192}$ \\  
  \hdashline
  GAIL & $\color{OliveGreen}{0.483} / \color{Orange}{0.429}$ & $\color{OliveGreen}{0.274} / \color{Orange}{0.185}$ & $\color{OliveGreen}{0.342} / \color{Orange}{0.257}$ \\ 
  \hdashline
  WFA-IRL(\textbf{ours}) & $\color{OliveGreen}{\mathbf{0.797}} / \color{Orange}{\mathbf{0.708}}$ & $\color{OliveGreen}{\mathbf{0.776}} / \color{Orange}{\mathbf{0.642}}$ & $\color{OliveGreen}{\mathbf{0.733}} / \color{Orange}{\mathbf{0.602}}$ \\ 
  \hdashline
  \makecell{WFA-IRL \\ w/o WFA}  & $\color{OliveGreen}{0.683} / \color{Orange}{0.514}$ & $\color{OliveGreen}{0.652} / \color{Orange}{0.488}$ & $\color{OliveGreen}{0.566} / \color{Orange}{0.390}$ \\ 
  \hline
  Expert & $\color{OliveGreen}{0.798} / \color{Orange}{0.718}$ & $\color{OliveGreen}{0.776} / \color{Orange}{0.668}$ & $\color{OliveGreen}{0.734} / \color{Orange}{0.639}$ \\ 
  \hdashline
  Optimal & $0.798$ & $0.776$ & $0.734$ \\ 
  \hdashline
  Random & $0.000$ & $0.000$ & $0.000$\\ 
  \hline
\end{tabular}
\end{table}

\subsection{Our method and baselines}
Our method WFA-IRL uses a neural network architecture to represent the cost function. For a detailed description, please refer to Appendix~\ref{sec:cost_representation}. We use the spectral learning algorithm in the Scikit-SpLearn toolbox \cite{Arrivault2017SpLearn} to learn the parameters $\psi$ of the WFA from the expert demonstrations $\calD$. In the implementation, we first compress the demonstration words $\sigma_{0:T_n}^n$ where consecutive identical symbols are removed. This greatly reduces the complexity of learning the WFA while keeping the symbol sequences unchanged. The hyperparameters for the spectral learning method are the \texttt{rank} of the automaton ($m$ in Theorem \ref{th:spectral}) and the sizes (\texttt{rows} and \texttt{cols}) of the prefix and suffix basis ($\calB = \prl{\calP, \calS}$ in Theorem \ref{th:spectral}), which determine the size of the Hankel matrix estimated empirically from demonstrations. 
The complexity is $O(n\times \texttt{rank} \times \texttt{rows} \times \texttt{cols})$ since we iterate through the ranks to find a minimal WFA and count the prefix and suffix frequencies for a given word.
The spectral method can learn almost perfectly with near zero loss in \eqref{eq:wfa_loss} for all tasks and the best hyperparameter configurations are shown in Table~\ref{tb:results}. We observe that larger WFA capacity is required to learn from suboptimal trajectories and thus more diverse words from $\color{Orange}{\calD_2}$.

\begin{figure}[t]
\centering
\begin{minipage}[t]{0.48\linewidth}
\centering
\includegraphics[width=\linewidth, height=14cm]{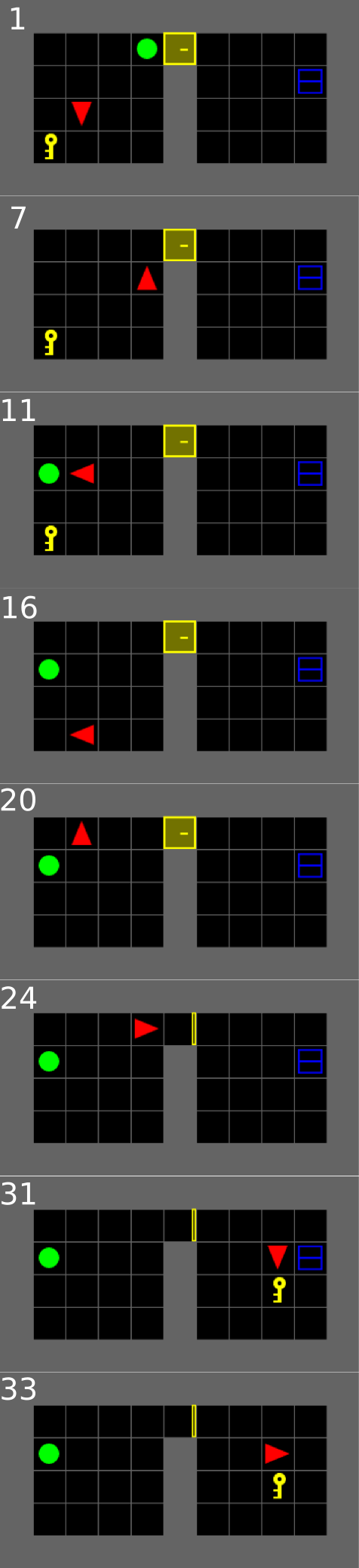}
\end{minipage}%
\hfill%
\begin{minipage}[t]{0.48\linewidth}
\centering
\includegraphics[width=\linewidth, height=14cm]{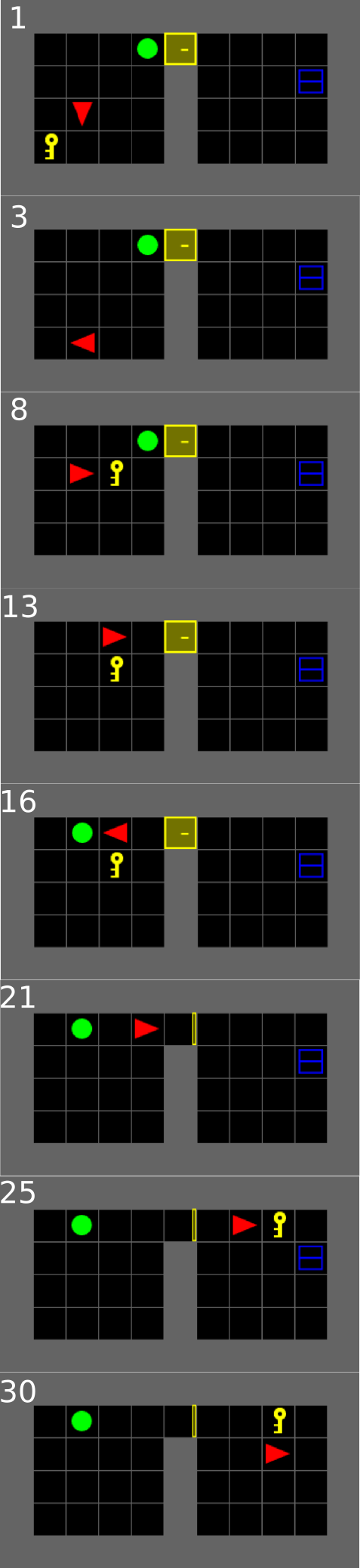}
\end{minipage}
\caption{Agent trajectory (left column) trained with $\color{OliveGreen}{\calD_1}$ and expert trajectory (right column) in task T3 during testing. The agent WFA only learns words that appear in demonstration $\color{OliveGreen}{\calD_1}$, which always moves $Ball$ away from $Door$ first before picking up $Key$. In testing it fails to recognize a lower cost trajectory of a different word sequence, where $Key$ is carried closer to $Door$ before moving away $Ball$.}
\label{fig:T3_different_words}
\end{figure}

As baselines, we first ablate the WFA component in our method to understand its effects. Instead of planning in the product WFA-MDP space and checking the WFA termination condition in \eqref{eq:Q_V}, the agent without WFA component simply plans in the original MDP and checks whether a goal state is achieved. Additionally, we compare our method with standard imitation learning and inverse reinforcement learning algorithms, including behavioral cloning (BC) \cite{Ross2010BC} and GAIL \cite{Ho2016GAIL}\footnote{The implementations are adapted from the imitation learning library \cite{Wang2020Imitation}}. The value/policy functions in these baselines follow our cost neural network architecture to fit the state-control input format and to compare fairly in representation power across methods. This includes the policy network in BC, the discriminator in GAIL, the policy and value networks in PPO \cite{Schulman2017PPO}, used as generator in GAIL. Only the size of the last fully-connected layer is modified depending on whether it is action or value prediction. GAIL is known to achieve stable performance in fixed horizon environments while the MiniGrid environments terminate as soon as the agent fulfills the tasks. We fix this issue by adding a virtual absorbing state as suggested in \cite{Kostrikov2019DAC} when training GAIL.

\begin{figure}[t]
\centering
\includegraphics[width=\linewidth]{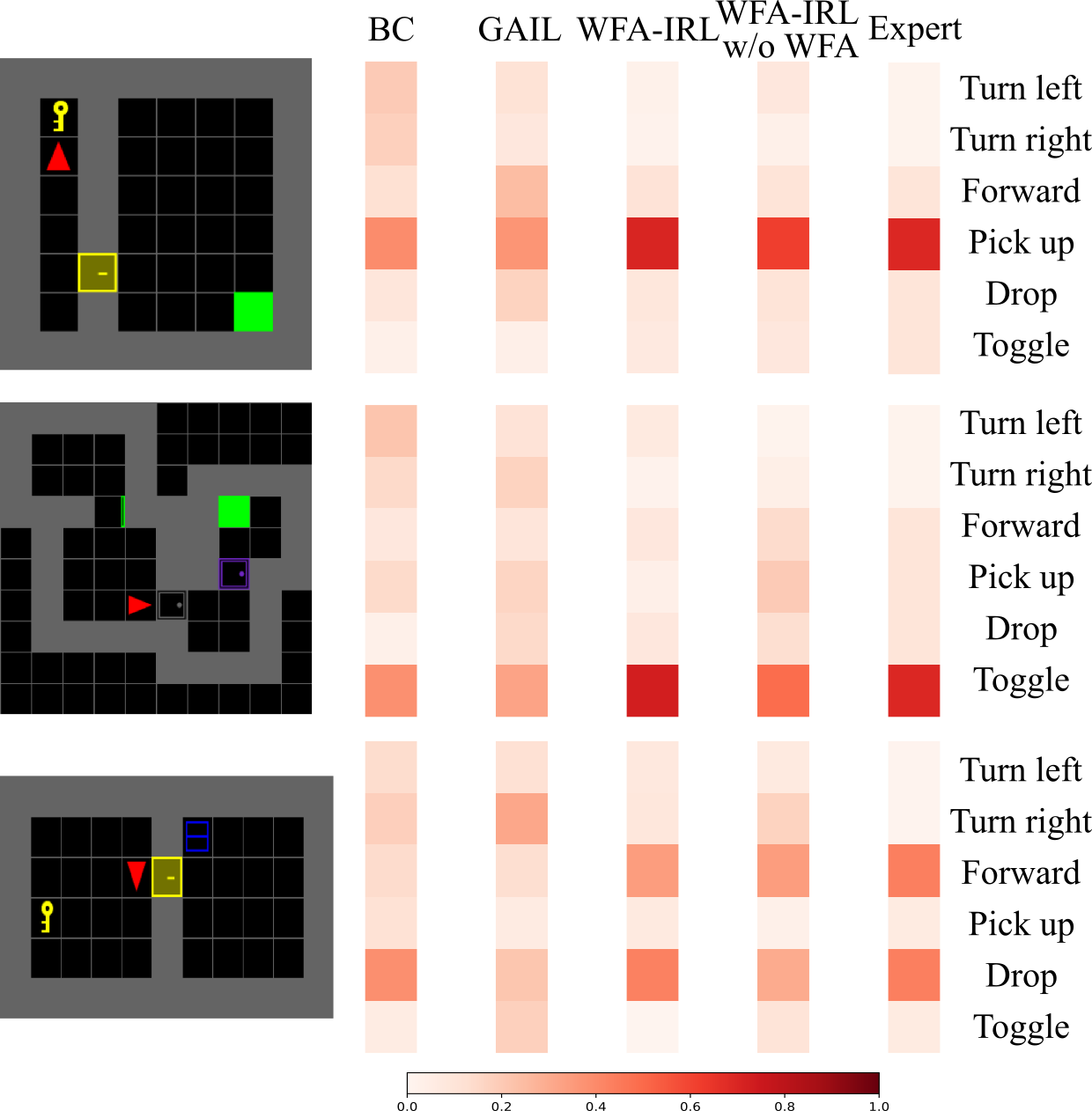}
\caption{Visualization of policy probabilities of each method trained on $\color{Orange}{\calD_2}$ at a critical state in T2. Our method shows a stronger preference towards controls (toggle door) that can make task progress.}
\label{fig:policy_comparison}
\end{figure}

\subsection{Results}
We report the average performance of each method in Table~\ref{tb:results} by testing on 64 new environment configurations generated randomly for each task. First, we observe that our method can achieve almost perfect performance when trained on $\color{OliveGreen}{\calD_1}$. This is expected since the learned WFA strictly chooses planned trajectories whose words would match the optimal behavior. Interestingly, the learned WFA could make the agent suboptimal if the optimal word in testing is not seen in training, as shown in Fig.~\ref{fig:T3_different_words}. Next, our method matches the expert performance well using either $\color{OliveGreen}{\calD_1}$ or $\color{Orange}{\calD_2}$ and outperforms BC and GAIL (even without WFA). This demonstrates that using planning to solve tasks that encode logical structures performs better than a reactive policy employed by BC. Moreover, the performance gap between our method and ours without WFA shows that learning logic specifications explicitly with a WFA can further improve the policy. On the other hand, we find the performance of GAIL is limited as PPO cannot easily generate successful samples similar to the demonstrations (notice that the random policy never succeeds) to improve the cost discriminator and, in turn, the generator itself. We visualize the agent policy in Fig.~\ref{fig:policy_comparison} and observe that our method has a stronger bias on controls that follow the learned logical sequences.

\section{Conclusion}
\label{sec:conclusion}
We present WFA-IRL which solves tasks with high-level reasoning and outperforms prior imitation learning and IRL methods that do not exploit logical structures from demonstrations. We show that cost functions learned via solving deterministic shortest path problems in the product WFA-MDP can generalize well in unseen environments and across demonstations of different optimality levels. 


\appendix
\section{Appendix}
\label{sec:appendix}


\begin{figure*}[ht]
\centering
\includegraphics[width=0.9\linewidth]{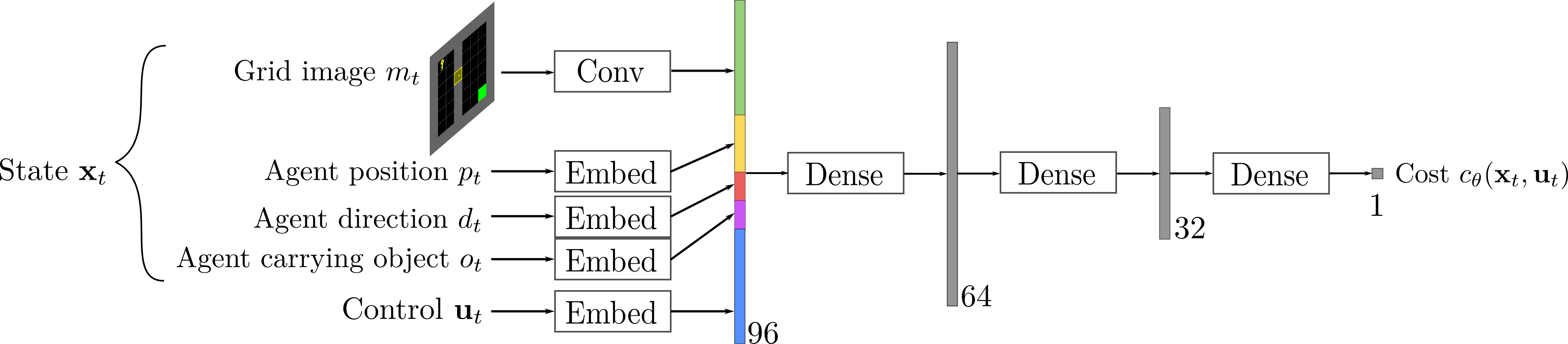}
\caption{Neural network architecture for the transition cost $c_\theta(\bfx_t, \bfu_t)$. The state $\bfx_t$ consists of the grid image $\bfm_t$, the agent position $\bfp_t$, direction $d_t$, object it is carrying $o_t$. The grid $m_t$ is fed through a convolutional neural network (Conv), while the discrete variables $\bfp_t$, $d_t$, $o_t$ and the control $\bfu_t$ are converted to embedding vectors (Embed) to provide latent representations for learning the cost function. The concatenated vector of Conv and Embed layer outputs is passed through three fully-connected layers (Dense) to obtain $c_\theta(\bfx_t, \bfu_t)$.}
\label{fig:cost_function}
\end{figure*}

\subsection{Neural network cost representation}
\label{sec:cost_representation}
We use a neural network, shown in Fig.~\ref{fig:cost_function} to learn a nonlinear cost function $c_\theta$, mapping from each state-control pair to a non-negative cost value. The cost neural network is separated into two parts. The first part is a feature extractor which processes each input type accordingly. The grid image is passed through a convolutional neural network, consisting of 3 stacks of convolution + ReLU layers with $\{16, 32, 64\}$ filters of size 2. The agent position, direction, object carried and control are discrete variables and are passed through embedding layers to produce high-dimensional feature vectors. The embedding dimensions are $\{128, 64, 64, 128\}$ respectively. The outputs from each feature extractor are flattened and concatenated to construct a latent vector representing the state-control pair in feature space. In the second part, a fully-connected neural network maps the latent vector to a scalar output for cost prediction. The 3 fully-connected layers have sizes $\{64, 32, 1\}$ with ReLU activation function. The cost neural network architecture is trained using Proposition~\ref{prop:chain_rule} in PyTorch \cite{Paszke2019Pytorch} with the Adam optimizer \cite{Kingma2014ADAM}.

\newpage
\balance
{\small
\bibliographystyle{cls/IEEEtran}
\bibliography{bib/ref.bib}
}

\end{document}